\documentclass{llncs}
\usepackage{xurl}
\usepackage{listings}
\usepackage{xcolor}
\usepackage{microtype}
\usepackage{inconsolata}
\usepackage{cite}
\usepackage{graphicx}
\usepackage{amsmath}
\usepackage{url}
\usepackage{amssymb}
\usepackage{booktabs}
\usepackage{colortbl}
\usepackage{array}
\usepackage[caption=false]{subfig}
\usepackage{placeins}
\usepackage[table]{xcolor}
\makeatletter
\renewcommand\section{\@startsection{section}{1}{\z@}%
  {-8pt plus -2pt minus -2pt}%
  {5pt}%
  {\normalfont\large\bfseries}}

\renewcommand\subsection{\@startsection{subsection}{2}{\z@}%
  {-6pt plus -1pt minus -1pt}%
  {4pt}%
  {\normalfont\normalsize\bfseries}}

\renewcommand\subsubsection{\@startsection{subsubsection}{3}{\z@}%
  {-5pt plus -1pt minus -1pt}%
  {3pt}%
  {\normalfont\normalsize\bfseries}}
\makeatother

\definecolor{lightgray}{gray}{0.95}
\definecolor{bestcolor}{RGB}{198,239,206}
\definecolor{besttext}{RGB}{0,100,0}
\definecolor{ourscolor}{RGB}{232,245,253}

\usepackage{multirow}
\usepackage{makecell}
\usepackage{algorithm}
\usepackage{algpseudocode}

\usepackage{float}
\floatname{algorithm}{Algorithm}

\lstdefinestyle{grayprompt}{
    backgroundcolor=\color{lightgray},
    basicstyle=\ttfamily\small,
    frame=single,
    breaklines=true,
    columns=fullflexible,
    keepspaces=true,
    showstringspaces=false
}

\title{MedStruct-S: A Benchmark for Key Discovery, Key-Conditioned QA and Semi-Structured Extraction from OCR Clinical Reports}
\author{Yingyun Li \and Yu Wang \and Haiyang Qian\thanks{Corresponding author: haiyang.qian@aistarfish.com}}
\authorrunning{Y. Li et al.}
\institute{AI Starfish, Hangzhou, China}

\begin{document}
\maketitle
\begin{abstract}
Semi-structured information extraction (IE) from OCR-derived clinical reports is crucial to efficiently reconstruct the patients' longitudinal medical history. In practice, this scenario commonly involves three tasks: (i)field-header (key) discovery, (ii)key-conditioned question-answering (QA), and (iii) end-to-end key--value pair extraction. However, existing evaluations often under-model two factors--heterogeneous and not-completely-known key representations and OCR-induced noise--making it difficult to assess model robustness in real-world settings.
We present MedStruct-S, a benchmark specifically designed to evaluate these tasks under unknown keys and OCR noise. MedStruct-S contains 3,582 annotated real-world clinical report pages. Using MedStruct-S, we benchmark two representative paradigms: encoder--only sequence labeling with post-processing and decoder-only structured generation—covering four encoder-only and five decoder-only models spanning 0.11B–103B parameters. Our results show that encoder-only models achieve best performance for non-null-value key-conditioned QA despite being substantially smaller than decoder-only models. Comparing models of similar order of magnitude, encoder-only models still perform better overall. Without controlling for model scale, the fine-tuned decoder only models deliver the strongest overall results. These processes illustrate that the benchmark provides a reliable and practical basis for selecting and comparing models across different semi-structured IE settings.

\end{abstract}
\keywords{Clinical Reports \and Benchmark \and OCR Noise\and Semi-Structured Extraction \and Key Discovery \and Key-Conditioned QA}

\section{Introduction}
Medical histories are usually not shared between different institutions, so clinicians in cross-provider care mostly rely on paper reports brought by patients to obtain prior information. Therefore, reliable semi-structured IE from clinical texts is important for obtaining medical histories. This process begins with OCR on images, followed by model-based extraction on the OCR-derived text.

\begin{figure*}[t]
    \centering
    \includegraphics[width=\linewidth]{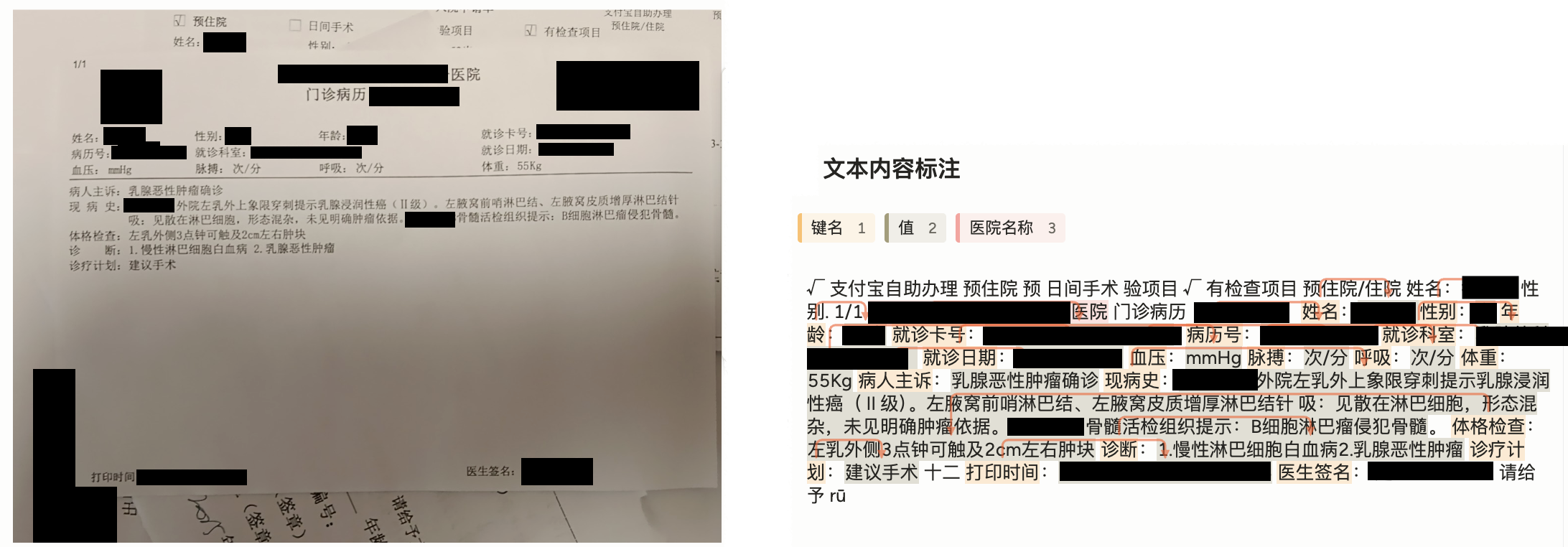}
    \caption{A clinical report page and its annotated key--value pairs}
    \label{fig:datasample_cn}
\end{figure*}

Clinical reports usually have an explicit layout: clinical concepts appear as field headers, followed by patient-specific text, separated by delimiters such as colons \cite{Fu_2020,jaume2019}. In this paper, field headers and patient-specific text are called keys and values, respectively. Fig.~\ref{fig:datasample_cn} shows a clinical report page with its annotated key--value pairs.

Most medical IE studies focus on clean electronic health record (EHR) text. However, when it comes to clinical text generated by OCR in real-world scenarios, encoder-only models are better at maintaining literal accuracy, whilst decoder-only models are more tolerant of OCR noise but are more prone to boundary drift. Existing benchmarks do not evaluate this setting, where keys are not predefined and may appear in diverse surface forms (i.e., aliases). To address this, we propose MedStruct-S, a benchmark for evaluating models on three tasks, as shown in Fig.~\ref{fig:benchmark_framework}.

We collect clinical reports from cancer patient care programs and run OCR on 3,582 pages. Through a 560-person-day annotation effort, the process results in MedStruct-S, a semi-structured benchmark built from OCR-derived clinical reports. For privacy and compliance, we also release a de-identified version, MedStruct-S (De-ID) is obtained by replacing sensitive information (\textsl{e.g.}, patient IDs, birth dates, and locations) with synthetic placeholders while preserving document structure and OCR noise. Its performance remains highly consistent with that of the original dataset under the same models and tasks.

To better characterize the impact of OCR noise, we evaluated extraction quality using both exact match (EM) and approximate match (AM), which distinguish literal boundary fidelity from semantic tolerance under noisy input.

The rest of this paper is organized as follows. Section~2 reviews related work. Section~3 presents MedStruct-S, including data construction, task definitions, and evaluation metrics. Section~4 reports the experimental setup, results, and analysis. Section~5 concludes and discusses limitations.

\section{Related Work}

Clinical IE from reports has traditionally been framed as a sequence labeling problem, with encoder-only models such as BERT \cite{cui-etal-2020-revisiting, cui2021pretraining} establishing strong baselines for named entity recognition (NER) and relation extraction on benchmarks such as CMeEE and CMeIE (F1 > 60\%--70\%). However, these models are mainly evaluated on closed-schema span matching, and their ability to perform schema-free key--value pairing without predefined keys remains unclear \cite{bhattacharyya-etal-2025-information}. Unified IE frameworks such as UIE \cite{lu2022unified} and InstructUIE \cite{wang2023instructuie} cast IE as structured generation or instruction following, but still assume task schemas or prompts rather than OCR-derived clinical reports with unknown keys.

Generative IE models include general-purpose LLMs such as Qwen3 \cite{Yang2025Qwen3TR} and the Qwen-2.5 series \cite{qwen2024technical}, which are pretrained on large-scale web data and exhibit strong instruction-following and zero-shot generalization capabilities. Medical-adapted LLMs, such as AntAngelMed \cite{antgroup2025antangelmed} and Baichuan\cite{baichuan2025m2}, are pretrained or fine-tuned using medical corpora to enhance medical reasoning and factual accuracy.

Current medical IE benchmarks mainly focus on clean clinical text. CBLUE \cite{zhang-etal-2022-cblue, zhang2021cblue}, including tasks such as CMeEE (NER) and CMeIE (relation extraction) \cite{guan2020cmeie}, is widely used but does not target robustness in real-world OCR scenarios. PromptCBLUE \cite{zhu2023promptcblue} reformulates classification and extraction as prompt-response tasks for instruction-following evaluation, but its performance remains sensitive to prompt wording. CHIP-CDN \cite{zhang-etal-2022-cblue} is limited to diagnosis normalization, IMCS-V2 \cite{chen2022imcs} lacks OCR-induced character noise, and English clinical IE benchmarks such as RadGraph \cite{jain2021radgraph} focus on digitally native radiology reports with predefined extraction schemas. Within the medical domain, EHRStruct \cite{Yang2025EHRStructAC} evaluates LLMs on structured electronic health record tasks, but its reliance on clean, digitally native EHR data leaves a gap with real-world deployment, where OCR-induced character-level and layout-level noise are common.

General document parsing benchmarks such as OmniDocBench \cite{ouyang2024omnidocbenchbenchmarkingdiversepdf} and Docopilot \cite{11094096} provide strong protocols for layout analysis and key information extraction, but are not tailored to clinical semantics, open-world key discovery in OCR-derived reports, or variable medical key aliases. MedStruct-S fills this gap by benchmarking semi-structured extraction from OCR-derived clinical reports under realistic OCR-induced noise and unknown key inventories (\textsl{i.e.}, keys are not provided \textit{a priori}), using EM and AM.

\section{The Proposed Benchmark}

\begin{figure}[!t]
  \centering
  \captionsetup{skip=3pt}
  \includegraphics[width=\textwidth,trim=0 20 0 0,clip]{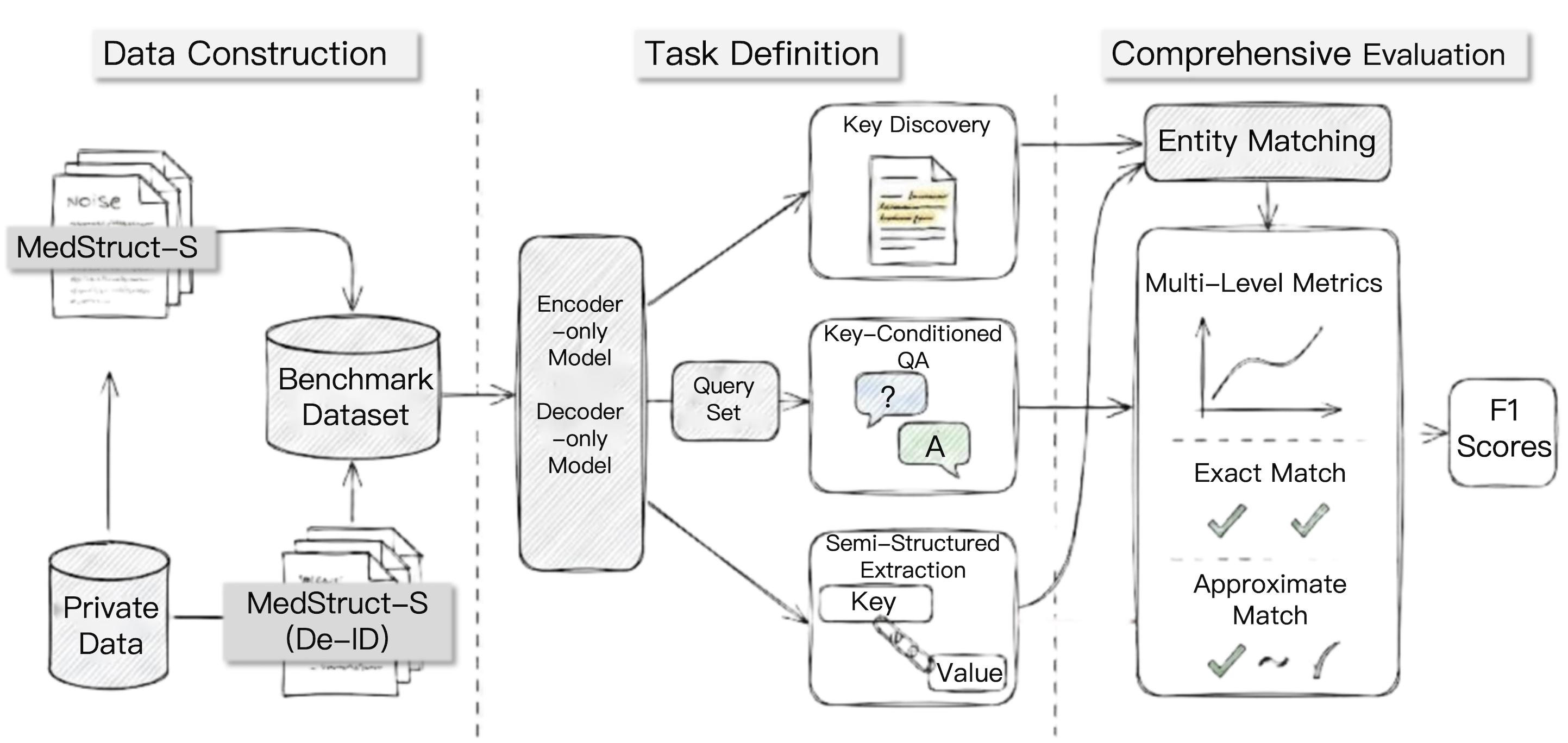}
  \caption{MedStruct-S: data corpus, task definitions, and evaluation metrics}
  \label{fig:benchmark_framework}
  \vspace{-4pt}
\end{figure}

As shown in Fig.~\ref{fig:benchmark_framework}, MedStruct-S provides an end-to-end benchmark for OCR-derived clinical reports, covering data construction, open-key extraction tasks, and evaluation.

\subsection{Data Construction}

\begin{figure*}[h]
    \centering
    \captionsetup{skip=2pt}
    \includegraphics[width=\textwidth]{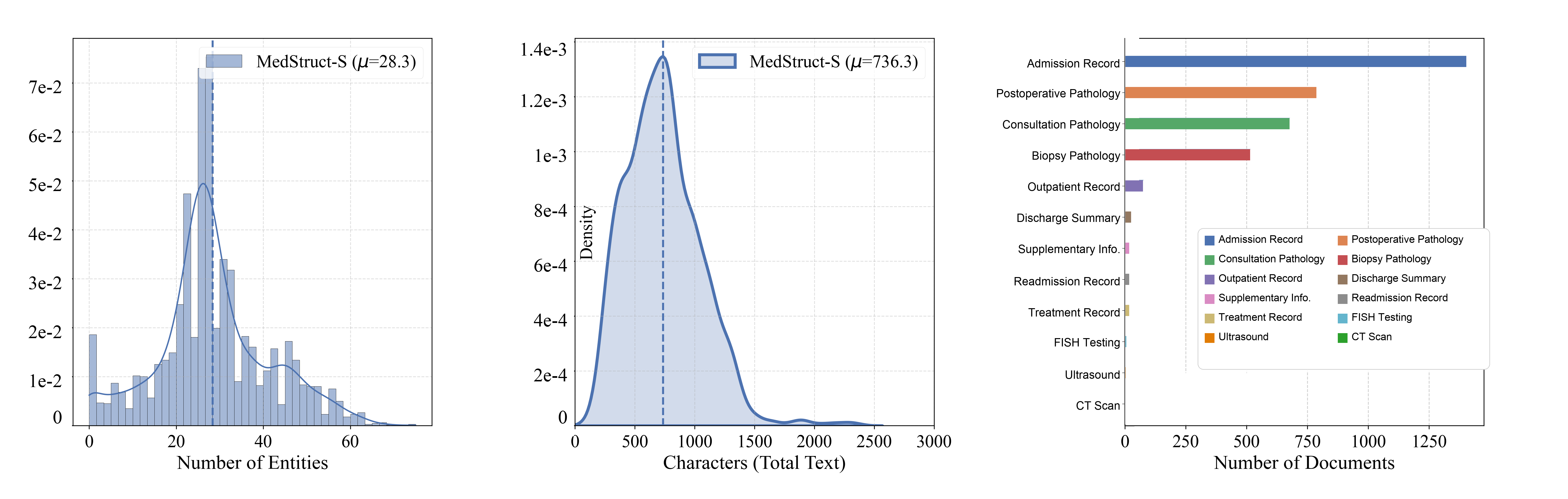}
    \caption{Distribution of categories and statistics on text length reported in MedStruct-S.}
    \label{fig:density_length_category}
\end{figure*}

The clinical reports in this dataset were collected with the patients' consent and converted to text using Baidu OCR \cite{baidu_ocr}. Annotation is performed by trained annotators using Label Studio \cite{LabelStudio}. The errors induced by OCR noise are not corrected during this process. To ensure annotation quality, 20\% of the samples were randomly selected for manual verification. MedStruct-S covers multiple categories of clinical reports (Fig.~\ref{fig:density_length_category}).

\begin{figure*}[h]
  \centering
  \includegraphics[width=\textwidth, trim=0 0 0 80, clip]{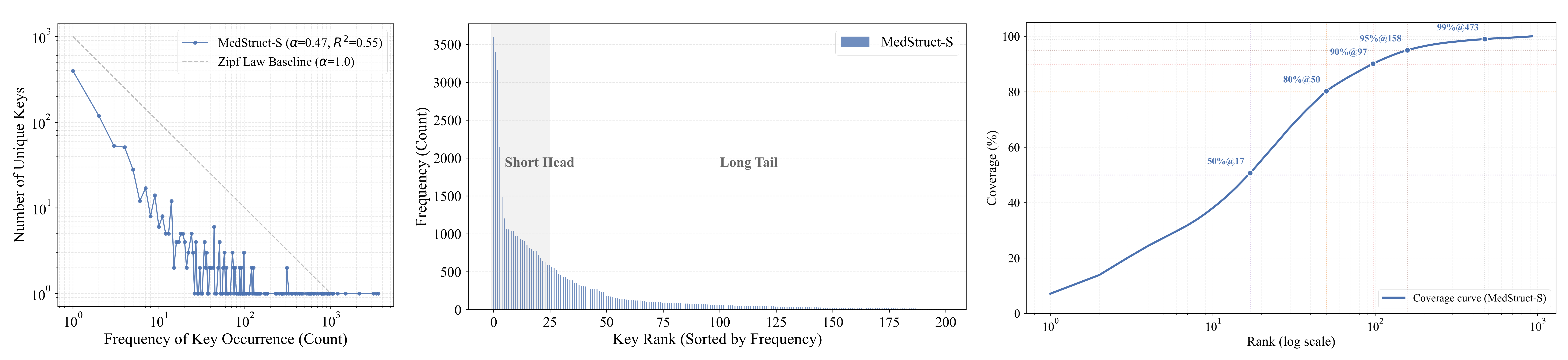}
  \caption{Key frequency distribution in MedStruct-S.}
  \label{fig:key_distribution}
\end{figure*}

We analyze the key distribution in MedStruct-S and observe a clear long-tail pattern shown in Fig.~\ref{fig:key_distribution}, with a small number of frequent keys and a large number of rare ones which are possibly unknown. Medical key extraction is therefore inherently an open-world problem, meaning that value extraction must be performed without assuming a complete predefined key inventory.

\begin{figure}[t]
  \centering
  \captionsetup{skip=2pt}
  \includegraphics[width=\linewidth]{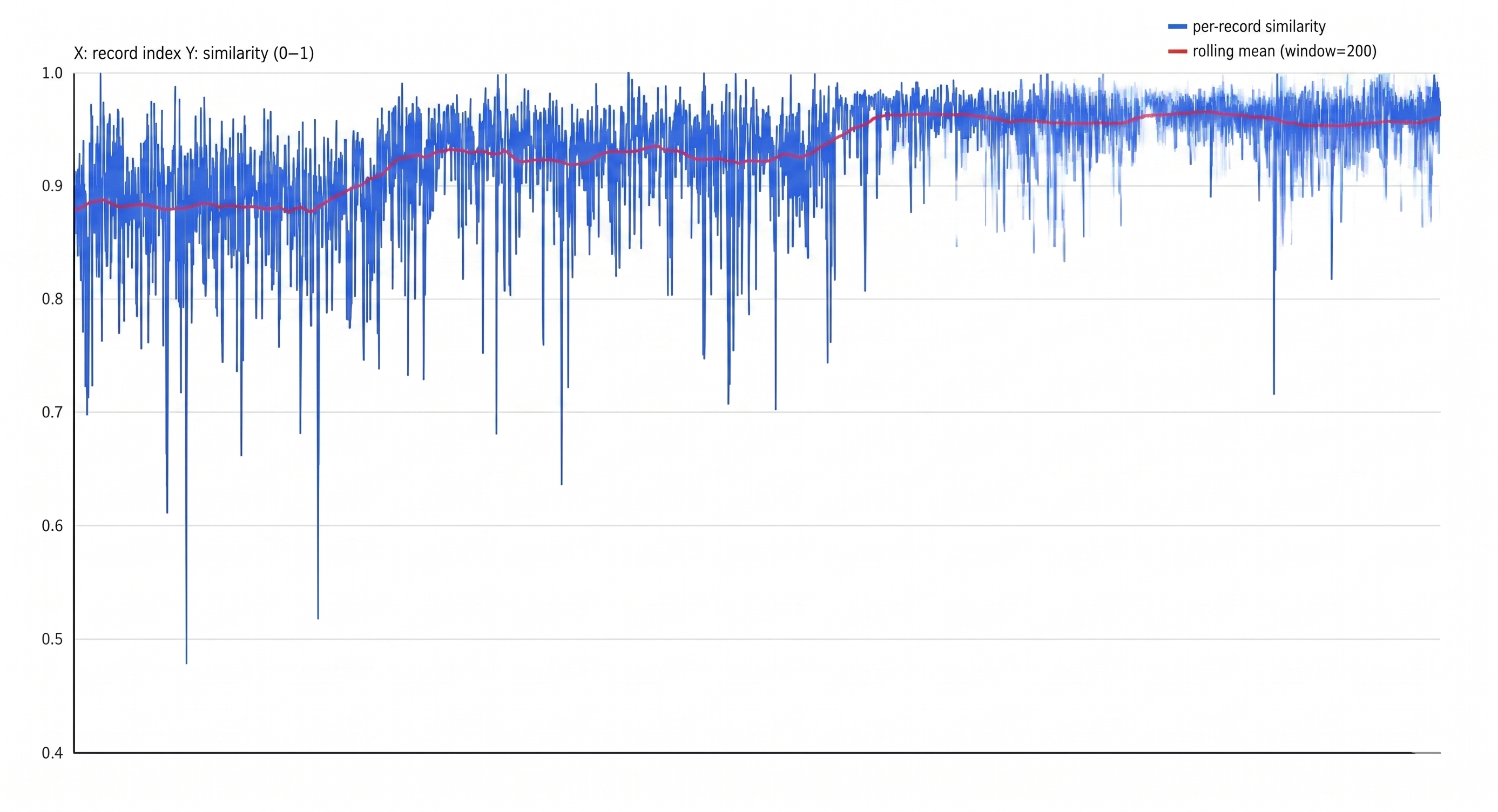}
  \caption{Page-level text similarity between the original dataset and de-identified variants.}
  \label{fig:deid_similarity}
\end{figure}

To support public release of the benchmark while ensuring patient privacy, MedStruct-S (De-ID) is constructed by replacing sensitive information with synthetic placeholders while preserving the original report structure and OCR noise characteristics. We quantify the fidelity of the de-identification process by measuring page-level text similarity between MedStruct-S and MedStruct-S (De-ID) (Fig.~\ref{fig:deid_similarity}). The consistently high page-level similarity indicates that de-identification introduces only minimal character-level changes while preserving the original OCR-induced noise patterns and structural layout of the reports.

\subsection{Task Definitions}

Let $p$ denote an OCR-derived clinical report page, $K$ the ground-truth key set, and $KV$ the ground-truth set of key--value pairs. We define three tasks. In \textbf{Task~1 (Key Discovery)}, the input is $p$ and the output is a predicted key set $\hat{K}$, identified without assuming a predefined key inventory. \textbf{Task~2 (Key-Conditioned QA)} takes $(p,k)$ as input, where $k$ is a queried key, and outputs the corresponding value $\hat{v}$; together, Tasks~1 and~2 form a two-stage pipeline for semi-structured IE. In \textbf{Task~3 (Semi-Structured Extraction)}, the input is $p$ and the output is a predicted set of key--value pairs $\widehat{KV}=\{(\hat{k},\hat{v})\}$, enabling end-to-end semi-structured extraction in a single step.

\subsection{Evaluation Metrics}

To quantify robustness to OCR-derived noise, we define a normalized similarity score $\phi(u, v) \in [0, 1]$ between two strings $u$ and $v$ based on the Levenshtein edit distance $d_{\mathrm{lev}}(u,v)$, where $|u|$ and $|v|$ denote string lengths:
\begin{equation*}
    \phi(u, v) = 1 - \frac{d_{\text{lev}}(u, v)}{\max(|u|, |v|)}.
\end{equation*}
We define $\operatorname{Norm}(\cdot)$ as a string normalization function that trims whitespace and converts characters to lowercase. We adopt a length-adaptive threshold $\tau(\ell)$ parameterized by the ground-truth string length $\ell$ as the acceptance criterion for approximate matching:
\begin{equation*}
\tau(\ell) =
\begin{cases}
0.8, & \text{if } \ell < 10, \\
0.8 + 0.01(\ell-10), & \text{if } 10 \le \ell \le 20, \\
0.9, & \text{if } \ell > 20.
\end{cases}
\end{equation*}

Before text evaluation, we compute span IoU between predictions and ground-truth entities, then perform greedy one-to-one matching using the highest unmatched IoU pairs.

For Task~1, a predicted key $\hat{k}$ is counted as a true positive if it is matched to a ground-truth key $k$ and satisfies either EM, where $\operatorname{Norm}(\hat{k})=\operatorname{Norm}(k)$, or AM, where $\phi(\hat{k},k)\ge\tau(|k|)$.

For Task~2, we report accuracy on both non-null-value samples ($QA^{\mathrm{nnv}}_{\mathrm{e/a}}$) and all samples ($QA_{\mathrm{e/a}}$), where $\mathrm{e}$ and $\mathrm{a}$ denote EM and AM, respectively. $QA^{\mathrm{nnv}}$ is computed only on instances whose gold value is non-\texttt{NULL}. A prediction is counted as correct when $\hat{v}$ matches $v$ under the specified strictness: $\operatorname{Norm}(\hat{v})=\operatorname{Norm}(v)$ for EM, or $\phi(\hat{v},v)\ge\tau(|v|)$ for AM. For overall accuracy, a prediction is also counted as correct when both the ground truth and the predicted value are \texttt{NULL}.

For Task~3, a predicted pair $(\hat{k}, \hat{v})$ is counted as a true positive if $\hat{k}$ is matched to a ground truth key $k$ by span IoU. We evaluate textual correctness under three strictness levels: $\mathrm{K_eV_e}$ requires exact matches for both key and value. $\mathrm{K_eV_a}$ demands an exact key match but permits an approximate value match $\phi(\hat{v},v)\ge\tau(|v|)$. Finally, $\mathrm{K_aV_a}$ allows approximate matching for both key and value, \textsl{i.e.}, $\phi(\hat{k},k)\ge\tau(|k|)$ and $\phi(\hat{v},v)\ge\tau(|v|)$.

Algorithm~\ref{alg:medstruct_pipeline} summarizes the MedStruct-S pipeline.

\begin{algorithm}[t]
\caption{MedStruct-S benchmark construction and evaluation pipeline}
\label{alg:medstruct_pipeline}
\begin{algorithmic}[1]
\Require OCR-derived clinical report pages $\mathcal{P}$
\Ensure Task instances and task-specific evaluation scores

\For{each page $p \in \mathcal{P}$}
    \State Annotate key spans and value spans without correcting OCR noise
    \State Construct ground-truth key set $K$ and key--value pair set $KV$
    \State Create Task~1 instance: $p \rightarrow \hat{K}$
    \For{each queried key $k$}
        \State Create Task~2 instance: $(p,k) \rightarrow \hat{v}$
    \EndFor
    \State Create Task~3 instance: $p \rightarrow \widehat{KV}$
\EndFor
\State Run model inference for Tasks~1--3
\State Perform greedy one-to-one span IoU matching and compute EM/AM
\end{algorithmic}
\end{algorithm}

\section{Experiment and Results}
\subsection{Models and Experiment Setup}

We evaluate encoder-only and decoder-only models on both MedStruct-S and MedStruct-S (De-ID), and report results for Tasks~1--3 on both datasets. We select four encoder-only models—M-BERT \cite{DBLP:journals/corr/abs-1810-04805}, RoBERTa \cite{liu2019robertarobustlyoptimizedbert}, MacBERT \cite{cui-etal-2020-revisiting}, and McBERT \cite{xu2020mcbertefficientlanguagepretraining}—to cover multilingual pretraining, Chinese-specific optimization, correction-aware pretraining, and medical domain adaptation. For decoder-only models, we include both general-purpose and medical-adapted LLMs across different size scales: Qwen3-0.6B and Qwen3-14B \cite{Yang2025Qwen3TR} to examine within-family scaling, Qwen2.5-32B \cite{qwen2024technical} and Baichuan-M2-32B to compare general and medically adapted 32B models, and AntAngelMed-103B \cite{antgroup2025antangelmed} as a large-scale medical reference. For all models and tasks, we use a fixed train/validation/test split of $0.81{:}0.09{:}0.1$ with seed 42.

\textbf{Encoder-only implementation.} We employ a BERT--BiLSTM--CRF sequence labeling model \cite{bert-bilstm-crf} with BIO tags to recover key spans (Task~1) and key/value spans (Task~3), trained for 30 epochs with AdamW (batch size 128, learning rate $2\times10^{-5}$, weight decay 0.03, warmup ratio 10\%). For Task~2, we fine-tune \texttt{BertForQuestionAnswering} \cite{wolf-etal-2020-transformers} for extractive QA and downsample no-answer cases to 0.4. For Task~3, each predicted key is linked to the nearest following value as a deterministic heuristic baseline rather than a learned key--value relation extractor. For long pages, we use overlapping segmentation during both training and inference (segment size 500, overlap 50), and merge segment predictions back to page level by original offsets after removing duplicate spans.

\textbf{Decoder-only implementation.} We evaluate decoder-only models with two-shot inference and LoRA fine-tuning, except AntAngelMed-103B, which is evaluated only with two-shot inference. We use a unified prompt library for both settings, with full prompt templates released in the project repository README, and adopt multilingual prompts, which perform slightly better in our setting \cite{tanwar-etal-2023-multilingual}. For LoRA fine-tuning, we use learning rate $5\times10^{-5}$, rank $r{=}8$, dropout 0.0, and apply LoRA to the attention projection modules (\texttt{q\_proj}, \texttt{k\_proj}, \texttt{v\_proj}, \texttt{o\_proj}). Inference uses vLLM with greedy decoding (temperature $=0$).

\subsection{Results and Analysis}

Tabs.~\ref{tab:results-raw} and \ref{tab:results-deid} report the results on MedStruct-S and MedStruct-S (De-ID), respectively. Tab.~\ref{tab:results-raw} summarizes the main results, and Tab.~\ref{tab:results-deid} shows a broadly similar pattern on the de-identified dataset. Unless otherwise specified, we discuss MedStruct-S and note differences for MedStruct-S (De-ID).

\begin{table*}[t]
  \centering
  \small
  \setlength{\tabcolsep}{3.5pt}
  \renewcommand{\arraystretch}{1.15}
  \definecolor{lightgray}{gray}{0.92}
  \caption{Model evaluation using MedStruct-S}
  \label{tab:results-raw}
  \resizebox{\textwidth}{!}{%
  \begin{tabular}{lllccccccccc}
    \toprule
    \multirow{2}{*}{Category} & \multirow{2}{*}{Model} & \multirow{2}{*}{Setup} & \multicolumn{2}{c}{Task 1: Key Discovery} & \multicolumn{4}{c}{Task 2: Key-Conditioned QA} & \multicolumn{3}{c}{Task 3: Semi-Structurized Extraction} \\
    \cmidrule(lr){4-5} \cmidrule(lr){6-9} \cmidrule(lr){10-12}
     &  & & $\mathrm{K}_{\mathrm{e}}$ &  $\mathrm{K}_{\mathrm{a}}$ &  $\mathrm{QA}_{\mathrm{e}}$ &  $\mathrm{QA}_{\mathrm{a}}$ &  $\mathrm{QA}^{\mathrm{nnv}}_{\mathrm{e}}$ & $\mathrm{QA}^{\mathrm{nnv}}_{\mathrm{a}}$ &  $\mathrm{K}_{\mathrm{e}}\mathrm{V}_{\mathrm{e}}$ &  $\mathrm{K}_{\mathrm{e}}\mathrm{V}_{\mathrm{a}}$ &  $\mathrm{K}_{\mathrm{a}}\mathrm{V}_{\mathrm{a}}$ \\
    \midrule
    
    \multirow{4}{*}{Encoder-only} & MBERT (0.18B) & Fine-tuning  & 0.7307 & 0.7314 & \textbf{0.8598} & \textbf{0.8689} & 0.7384 & 0.7966 &  0.6450 & 0.6996 & 0.7001 \\
    &RoBERTa-wwm (0.11B) & Fine-tuning  & 0.7348 &  0.7352 & 0.7482 & 0.7577 & 0.7451 & 0.8053 & 0.6359 &  0.6924 & 0.6926 \\
    &MacBERT (0.11B) & Fine-tuning  & 0.7338 &  0.7347 & 0.7885 & 0.7982 & \textbf{0.7468} & \textbf{0.8084} & 0.6458 & 0.7052 & \textbf{0.7061} \\
    &McBERT (0.11B) & Fine-tuning  & \textbf{0.7368} & \textbf{0.7372} & 0.7093 & 0.7185 & 0.7432 & 0.8017 & \textbf{0.6483} &  \textbf{0.7053}  &  0.7057 \\
    \midrule
    
    \multirow{9}{*}{Decoder-only} & \multirow{2}{*}{Qwen3-0.6B} & Two-shot  & 0.0684 & 0.0751 & 0.8635 & 0.8657 & 0.1878 & 0.2013 & 0.3347 & 0.3657 & 0.3673 \\
    & & \cellcolor{lightgray}LoRA & \cellcolor{lightgray}0.7956 & \cellcolor{lightgray}0.7977 & \cellcolor{lightgray}0.9261 & \cellcolor{lightgray}0.9391 & \cellcolor{lightgray}0.6236 & \cellcolor{lightgray}0.7068 & \cellcolor{lightgray}0.4931 & \cellcolor{lightgray}0.5452 & \cellcolor{lightgray}0.5470 \\
    \cmidrule{2-12}
    
    & \multirow{2}{*}{Qwen3-14B} & Two-shot & 0.4865 & 0.4890 & 0.9236 & 0.9364 & 0.6076 & 0.6893 & 0.4532 & 0.5187 & 0.5213 \\
    & & \cellcolor{lightgray}LoRA & \cellcolor{lightgray}0.8473 & \cellcolor{lightgray}0.8488 & \cellcolor{lightgray}0.9281 & \cellcolor{lightgray}\textbf{0.9409} & \cellcolor{lightgray}0.6363 & \cellcolor{lightgray}\textbf{0.7181} & \cellcolor{lightgray}0.6712 & \cellcolor{lightgray}0.7753 & \cellcolor{lightgray}0.7763 \\
    \cmidrule{2-12}
    
    & \multirow{2}{*}{Qwen2.5-32B-Instruct} & Two-shot & 0.5614 & 0.5632 & 0.9162 & 0.9220 & 0.5593 & 0.5966 & 0.4924 & 0.5516 & 0.5537 \\
    & & \cellcolor{lightgray}LoRA & \cellcolor{lightgray}0.8564 & \cellcolor{lightgray}0.8582 & \cellcolor{lightgray}0.9288 & \cellcolor{lightgray}0.9404 & \cellcolor{lightgray}0.6403 & \cellcolor{lightgray}0.7147 & \cellcolor{lightgray}0.6867 & \cellcolor{lightgray}0.7864 & \cellcolor{lightgray}0.7882 \\
    \cmidrule{2-12}
    
    & \multirow{2}{*}{Baichuan-M2-32B} & Two-shot & 0.4412 & 0.4433 & 0.9217 & 0.9329 & 0.5940 & 0.6659 & 0.4848 & 0.5580 & 0.5604 \\
    & & \cellcolor{lightgray}LoRA & \cellcolor{lightgray}\textbf{0.8624} & \cellcolor{lightgray}\textbf{0.8640} & \cellcolor{lightgray}\textbf{0.9290} & \cellcolor{lightgray}0.9400 & \cellcolor{lightgray}\textbf{0.6422} & \cellcolor{lightgray}0.7122 & \cellcolor{lightgray}\textbf{0.6884} & \cellcolor{lightgray}\textbf{0.7869} & \cellcolor{lightgray}\textbf{0.7884} \\
    \cmidrule{2-12}
    
    & AntAngelMed-103B & Two-shot & 0.6466 & 0.6494 & 0.9201 & 0.9295 & 0.5774 & 0.6377 & 0.4544 & 0.5147 & 0.5168 \\
    \bottomrule
  \end{tabular}%
  }
\end{table*}
\begin{table*}[t]
  \centering
  \small
  \setlength{\tabcolsep}{3.5pt}
  \renewcommand{\arraystretch}{1.15}
  \definecolor{lightgray}{gray}{0.92}
  \caption{Model evaluation using MedStruct-S(De-ID)}
  \label{tab:results-deid}
  \resizebox{\textwidth}{!}{%
  \begin{tabular}{lllccccccccc}
    \toprule
    \multirow{2}{*}{Category} &\multirow{2}{*}{Model} & \multirow{2}{*}{Setup} & \multicolumn{2}{c}{Task 1: Key Discovery} & \multicolumn{4}{c}{Task 2: Key-Conditioned QA} & \multicolumn{3}{c}{Task 3: Semi-Structurized Extraction} \\
    \cmidrule(lr){4-5} \cmidrule(lr){6-9} \cmidrule(l){10-12}
     & &  &  $\mathrm{K}_{\mathrm{e}}$ &  $\mathrm{K}_{\mathrm{a}}$ &  $\mathrm{QA}_{\mathrm{e}}$ &  $\mathrm{QA}_{\mathrm{a}}$ &  $\mathrm{QA}^{\mathrm{nnv}}_{\mathrm{e}}$ &  $\mathrm{QA}^{\mathrm{nnv}}_{\mathrm{a}}$ &  $\mathrm{K}_{\mathrm{e}}\mathrm{V}_{\mathrm{e}}$ &  $\mathrm{K}_{\mathrm{e}}\mathrm{V}_{\mathrm{a}}$ &  $\mathrm{K}_{\mathrm{a}}\mathrm{V}_{\mathrm{a}}$ \\
    \midrule
    
    \multirow{4}{*}{Encoder-only} &MBERT (0.18B) & Fine-tuning  & 0.7412  & 0.7417  & \textbf{0.8495} & \textbf{0.8556} & 0.7429 & 0.7817 & 0.6531 & 0.7025 &  0.7028 \\
    &RoBERTa-wwm (0.11B) & Fine-tuning  &  0.7407 &  0.7411  & 0.8153 & 0.8222 & 0.7463 & \textbf{0.7902} &  0.6492 & 0.7034 & 0.7034 \\
    &MacBERT (0.11B) & Fine-tuning  & \textbf{0.7466} & \textbf{0.7466} & 0.8058 & 0.8124 & \textbf{0.7468} & 0.7890 & \textbf{0.6621} & \textbf{0.7139} & \textbf{0.7139} \\
    &McBERT (0.11B) & Fine-tuning  &   0.7413 &   0.7417 & 0.7219 & 0.7286 & 0.7345 & 0.7772 & 0.6541 &  0.7071 & 0.7075 \\
    \midrule
    
    \multirow{9}{*}{Decoder-only} & \multirow{2}{*}{Qwen3-0.6B} & Two-shot  & 0.0784 & 0.0811 & 0.8664 & 0.8682 & 0.2224 & 0.2340 & 0.3372 & 0.3671 & 0.3684 \\
    & & \cellcolor{lightgray}LoRA & \cellcolor{lightgray}0.8006 & \cellcolor{lightgray}0.8027 & \cellcolor{lightgray}0.9269 & \cellcolor{lightgray}0.9383 & \cellcolor{lightgray}0.6256 & \cellcolor{lightgray}0.6989 & \cellcolor{lightgray}0.4951 & \cellcolor{lightgray}0.5543 & \cellcolor{lightgray}0.5557 \\
    \cmidrule{2-12}
    
    & \multirow{2}{*}{Qwen3-14B} & Two-shot & 0.4971 & 0.4995 & 0.9240 & 0.9358 & 0.6073 & 0.6831 & 0.4526 & 0.5152 & 0.5178 \\
    & & \cellcolor{lightgray}LoRA & \cellcolor{lightgray}0.8428 & \cellcolor{lightgray}0.8441 & \cellcolor{lightgray}0.9269 & \cellcolor{lightgray}\textbf{0.9395} & \cellcolor{lightgray}0.6256 & \cellcolor{lightgray}\textbf{0.7065} & \cellcolor{lightgray}0.6576 & \cellcolor{lightgray}0.7576 & \cellcolor{lightgray}0.7588 \\
    \cmidrule{2-12}
    
    & \multirow{2}{*}{Qwen2.5-32B-Instruct} & Two-shot & 0.5564 & 0.5584 & 0.9162 & 0.9201 & 0.5565 & 0.5819 & 0.4874 & 0.5403 & 0.5423 \\
    & & \cellcolor{lightgray}LoRA & \cellcolor{lightgray}0.8592 & \cellcolor{lightgray}0.8610 & \cellcolor{lightgray}0.9280 & \cellcolor{lightgray}0.9390 & \cellcolor{lightgray}0.6326 & \cellcolor{lightgray}0.7031 & \cellcolor{lightgray}\textbf{0.6731} & \cellcolor{lightgray}\textbf{0.7635} & \cellcolor{lightgray}\textbf{0.7648} \\
    \cmidrule{2-12}
    
    & \multirow{2}{*}{Baichuan-M2-32B} & Two-shot & 0.5235 & 0.5260 & 0.9219 & 0.9325 & 0.5926 & 0.6606 & 0.4835 & 0.5528 & 0.5556 \\
    & & \cellcolor{lightgray}LoRA & \cellcolor{lightgray}\textbf{0.8621} & \cellcolor{lightgray}\textbf{0.8637} & \cellcolor{lightgray}\textbf{0.9287} & \cellcolor{lightgray}0.9393 & \cellcolor{lightgray}\textbf{0.6372} & \cellcolor{lightgray}0.7048 & \cellcolor{lightgray}0.6669 & \cellcolor{lightgray}0.7619 & \cellcolor{lightgray}0.7632 \\
    \cmidrule{2-12}
    
    & AntAngelMed-103B & Two-shot & 0.6378 & 0.6407 & 0.9217 & 0.9303 & 0.5813 & 0.6363 & 0.4533 & 0.5090 & 0.5126 \\
    \bottomrule
  \end{tabular}%
}
\end{table*}

(i) Task~1 (Key Discovery). Two-shot decoder-only models perform poorly on key discovery, especially at small scale (e.g., Qwen3-0.6B, $\mathrm{K}_{\mathrm{e}}=0.0684$). LoRA substantially improves this behavior (to $0.7956$), and the best decoder-only Task~1 results are achieved by Baichuan-M2-32B (LoRA): $\mathrm{K}_{\mathrm{e}}/\mathrm{K}_{\mathrm{a}}=0.8624/0.8640$ on MedStruct-S and $0.8621/0.8637$ on MedStruct-S (De-ID).

(ii) Task~2 (Key-Conditioned QA). Decoder-only models achieve high accuracy on the overall evaluation set, but their performance drops more on \textit{non-null} samples. For instance, Qwen3-14B (LoRA) reaches $\mathrm{QA}_{\mathrm{e}}=0.9281$ overall but decreases to $\mathrm{QA}^{\mathrm{nnv}}_{\mathrm{e}}=0.6363$, whereas MacBERT attains a higher $\mathrm{QA}^{\mathrm{nnv}}_{\mathrm{e}}=0.7468$. Two error patterns help explain this gap in long, noisy OCR-derived text: decoder-only models may default to \texttt{NULL} when key/value evidence is difficult to localize, and they are more prone to \textit{boundary drift}, for example, truncating ''311 Area" to''311". Because encoder-only models predict spans directly, they provide more reliable localization on non-null cases.

(iii) Task~3 (Semi-Structured Extraction). For fine-tuned decoder-only models, AM yields larger gains than EM. On MedStruct-S, the best decoder-only Task~3 result is achieved by Baichuan-M2-32B (LoRA) ($\mathrm{K}_{\mathrm{e}}\mathrm{V}_{\mathrm{e}}=0.6884$, $\mathrm{K}_{\mathrm{a}}\mathrm{V}_{\mathrm{a}}=0.7884$); on MedStruct-S (De-ID), the best result is achieved by Qwen2.5-32B-Instruct (LoRA) ($\mathrm{K}_{\mathrm{e}}\mathrm{V}_{\mathrm{e}}=0.6731$, $\mathrm{K}_{\mathrm{a}}\mathrm{V}_{\mathrm{a}}=0.7648$).

Overall, MedStruct-S makes the trade-off between literal fidelity (EM) and semantic tolerance (AM) explicit, and the similar results on MedStruct-S (De-ID) suggest that these patterns are robust across the original and de-identified settings. For decoder-only models, scale matters, but model family and post-training also matter. In Task~3, the best decoder-only result is achieved by Baichuan-M2-32B (LoRA) on MedStruct-S ($\mathrm{K}_{\mathrm{a}}\mathrm{V}_{\mathrm{a}}=0.7884$) and by Qwen2.5-32B-Instruct (LoRA) on MedStruct-S (De-ID) ($\mathrm{K}_{\mathrm{a}}\mathrm{V}_{\mathrm{a}}=0.7648$).

\section{Conclusion and Limitations}

We presented MedStruct-S, a benchmark for semi-structured extraction from OCR-derived clinical reports. It contains 3,582 annotated real-world pages and covers three settings: key discovery, key-conditioned QA, and end-to-end key--value extraction, under realistic OCR noise and key variability, with both EM and AM metrics. Across MedStruct-S and its de-identified version, EM and AM reveal different robustness profiles, and end-to-end extraction remains particularly sensitive to format compliance and boundary accuracy, especially for smaller decoder-only models. MedStruct-S therefore provides a practical basis for model selection and paradigm comparison in OCR-derived clinical IE.

Task~3 is not fully symmetric: encoder-only models perform span annotation followed by a deterministic nearest-neighbor pairing heuristic, whereas decoder-only models generate structured output directly. MedStruct-S is currently Chinese-only, and both key realizations and OCR error patterns may differ across languages and scripts. Although the benchmark contains 3,582 annotated pages, it does not yet cover the full diversity of clinical reports across institutions and layouts. Future work will extend MedStruct-S to English and multilingual settings and broaden its coverage of institutions, report types, and layouts.

\section*{Acknowledgements}
We would like to thank our colleagues QIN Ying, JI Xiaosen, and WU Guifeng. In particular, we thank QIN Ying for identifying and proposing the problem of semi-structured information extraction from OCR-derived clinical reports, a line of work with broad implications for the healthcare industry. We also thank JI Xiaosen and WU Guifeng for providing the real-world data that made this research possible.

\section*{Code and Data Availability}
Data, code, and prompt templates are available at \url{https://github.com/AI-Starfish-Research/MedStruct-S}.

\bibliographystyle{splncs04}
\bibliography{reference}

\end{document}